\documentclass[conference]{IEEEtran}
\IEEEoverridecommandlockouts
\usepackage{cite}
\usepackage{amsmath,amssymb,amsfonts}
\usepackage{algorithmic}
\usepackage{textcomp}
\usepackage{xcolor}
\usepackage{booktabs}
\usepackage{multirow}
\usepackage{caption}
\usepackage{subcaption}  
\usepackage{threeparttable} 
\usepackage{makecell}
\usepackage{array}
\usepackage{enumitem}
\usepackage{pgfgantt}
\usepackage{graphicx}
\usepackage{float}
\usepackage{longtable}
\usepackage{url} 
\usepackage{tabularx,adjustbox,pifont}
\usepackage{placeins}
\usepackage[
  colorlinks=true,
  urlcolor=blue,
  citecolor=black,
  linkcolor=black
]{hyperref}
\def\BibTeX{{\rm B\kern-.05em{\sc i\kern-.025em b}\kern-.08em
    T\kern-.1667em\lower.7ex\hbox{E}\kern-.125emX}}

\begin{document}

\title{From Baseline to Follow-Up: Counterfactual Spine DXA Image Synthesis in UK Biobank Using a Causal Hierarchical Variational Autoencoder\\
% {\footnotesize \textsuperscript{*}Note: Sub-titles are not captured for https://ieeexplore.ieee.org  and
}

\author{
\IEEEauthorblockN{Yilin Zhang}
\IEEEauthorblockA{\textit{School of Electronics and Computer Science} \\
\textit{University of Southampton} \\
Southampton, UK \\
Yilin.Zhang@soton.ac.uk}
\and
\IEEEauthorblockN{Nicholas~C.~Harvey}
\IEEEauthorblockA{\textit{MRC Lifecourse Epidemiology Centre} \\
\textit{University of Southampton, Southampton General Hospital} \\
Southampton, UK \\
nch@mrc.soton.ac.uk}
\and
\IEEEauthorblockN{Nicholas~R.~Fuggle}
\IEEEauthorblockA{\textit{MRC Lifecourse Epidemiology Centre} \\
\textit{University of Southampton, Southampton General Hospital} \\
Southampton, UK \\
nrf@mrc.soton.ac.uk}
\and
\IEEEauthorblockN{Rahman~Attar\IEEEauthorrefmark{1}, \textit{Senior Member, IEEE}}
\IEEEauthorblockA{\textit{School of Electronics and Computer Science} \\
\textit{University of Southampton} \\
Southampton, UK \\
r.attar@soton.ac.uk \\
\IEEEauthorrefmark{1}Corresponding author}
}

\maketitle

\begin{abstract}
Dual-energy X-ray absorptiometry (DXA) is widely used for large-scale skeletal assessment, yet learning controllable and interpretable factor-specific anatomical variation remains challenging. We propose a metadata-conditioned causal hierarchical variational autoencoder (CHVAE) for causally consistent generation of anteroposterior (AP) spine DXA images from the UK Biobank (UKB). The model is trained on 3,743 raw AP spine scans from the first imaging visit and conditioned on basic participant attributes and lumbar morphometry. Causal consistency is evaluated in a baseline-to-follow-up setting using abduction--action--prediction (AAP): latent variables are abducted from baseline images, age is intervened to the repeat-imaging value, and the resulting counterfactual follow-up morphometry is compared with observed repeat-imaging measurements. Results show strong absolute-level agreement for key vertebral morphometry variables under age intervention, supporting intervention-aligned synthesis of anatomically plausible DXA images.
\end{abstract}


\section{Introduction}
Population ageing motivates quantitative approaches to characterize how anatomical structures evolve across the adult lifespan \cite{iburg2023burden}. Large-scale imaging cohorts such as UK Biobank (UKB) provide longitudinal data at population scale, including dual-energy X-ray absorptiometry (DXA) and a repeat-imaging arm that enables within-subject follow-up analysis \cite{littlejohns2020uk, bycroft2018uk}. DXA offers rapid, low-dose assessment of musculoskeletal structure. Spine DXA further supports vertebral morphometry and vertebral fracture assessment (VFA) via geometric measurements and semi-automated analysis pipelines \cite{dr2000morphometric, roberts2010detection, lems2021vertebral}.

Automated vertebral shape extraction from DXA has been studied using statistical shape/appearance models \cite{roberts2005vertebral, roberts2010detection}, and vertebral dimensions vary systematically with age, sex and body size \cite{junno2015age, hipp2022definition}. However, modelling subject-specific longitudinal morphological change remains challenging because repeat scans are limited and separated by multi-year intervals \cite{littlejohns2020uk, sudlow2015uk, dao2024conditional}. These limitations motivate generative approaches that map a baseline scan to a plausible follow-up while supporting controlled ``what-if'' analyses of specific factors such as age.

Causal generative modelling provides a principled route to separate intervention effects from confounding \cite{kocaoglu2017causalgan, yang2021causalvae, sanchez2022diffusion}. Following the Deep Structural Causal Model (DSCM) paradigm, metadata are represented with an explicit structural causal model (SCM) and counterfactual queries are answered via abduction–action–prediction (AAP), reusing exogenous noise to preserve subject identity under interventions \cite{pearl2009causality, pawlowski2020deep}. To improve the expressivity of the image component for multi-scale medical structure, a two-level hierarchical VAE (HVAE) is adopted, which has been shown to better capture multi-scale variability than shallow latent formulations \cite{dorent2023unified, vahdat2020nvae}.

This work studies baseline-to-follow-up AP spine DXA generation using an HVAE observation model coupled with an SCM over metadata. The effect of an age intervention $\mathrm{do}(\mathrm{age})$ on interpretable lumbar morphometry (L1--L4 width, height and derived area) is evaluated by generating follow-up predictions from baseline observations under $\mathrm{do}(\mathrm{age}=\mathrm{age}_{\mathrm{inst3}})$ and comparing them against repeat-imaging measurements. The main contributions of this work are summarized as follows:

\begin{itemize}
    \item A baseline-to-follow-up counterfactual setting for AP spine DXA is formulated using repeat-imaging subjects, generating follow-up via AAP under $\mathrm{do}(\mathrm{age}=\mathrm{age}_{\mathrm{inst3}})$.
    \item A two-level HVAE observation model is integrated into a DSCM framework to improve reconstruction fidelity and intervention-aligned counterfactual generation while preserving subject identity.
    \item Longitudinal validation is provided using absolute-level agreement between counterfactual follow-up predictions and observed follow-up morphometry, alongside change-based evaluation.
    
\end{itemize}

The code of this study is available at: \url{https://github.com/YilinZhang00/CHVAE}

\section{Materials and Methods}
\subsection{Dataset}
Data were obtained from the UKB under application 700191. This study focuses on AP spine DXA from the first imaging visit (recorded as instance~2 in UKB). AP spine DXA is clinically relevant for osteoporosis assessment because it captures the central axial skeleton and provides consistent L1--L4 vertebral morphometry measurements (average height, average width, and area) used in this work. In addition, spine DXA-derived endpoints are sensitive to metabolic bone changes and are therefore informative for screening and monitoring.

A total of 4,900 participants were initially selected based on the availability of instance~2 variables (age at DXA, sex, standing height, weight, and L1--L4 measures). After data cleaning and excluding incomplete records, 3,743 samples remained. The resulting cohort is used for both causal discovery and counterfactual image generation. A summary of variables is reported in Table~\ref{tab:metadata_summary}.

\begin{table}[ht]
\centering
\begin{threeparttable}
\caption{Summary of structured metadata variables (UKB DXA instance 2).}
\label{tab:metadata_summary}
\begin{tabular}{llll}
\hline
\textbf{Variable} & \textbf{Unit} & \textbf{Type} & \textbf{Distribution(mean $\pm$ std)} \\
\hline
Age at DXA & year & Numerical & $61.99 \pm 7.54$ \\
Standing height & cm & Numerical & $169.91 \pm 9.34$ \\
Weight & kg & Numerical & $76.32 \pm 15.05$ \\
L1--L4 average height & cm & Numerical & $13.86 \pm 0.90$ \\
L1--L4 average width & cm & Numerical & $4.34 \pm 0.41$ \\
L1--L4 area & cm$^2$ & Numerical & $60.43 \pm 8.35$ \\
Sex & -- & Categorical &
\begin{tabular}[t]{@{}l@{}}
Female: 1899 (50.73\%) \\
Male: 1844 (49.27\%)
\end{tabular} \\
\hline
\end{tabular}
\begin{tablenotes}
\footnotesize
\item[1] "--" indicates no unit for this variable.
%\item[2] Numerical variables are reported as mean $\pm$ standard deviation (n=3,743).
%\item[2] Sex is encoded as 0 for Female and 1 for Male.
\end{tablenotes}
\end{threeparttable}
\end{table}

\subsection{Methods}
\subsubsection{\textbf{Causal Discovery}}
\label{sec:causal_discovery}

A directed acyclic graph (DAG) capturing physiologically plausible relationships among the structured covariates is estimated using a hybrid constraint-based and functional-modelling procedure. Domain priors restrict the search space: \textit{Age} and \textit{Sex} are treated as roots, \textit{L1--L4 area} as a sink, and only directions from anthropometrics to vertebral geometry are allowed, i.e.,
$\mathcal{B}=\{\text{Age},\text{Sex},\text{Standing height},\text{Weight}\}\rightarrow
\mathcal{V}=\{\text{L1--L4 avg height},\text{L1--L4 avg width},\text{L1--L4 area}\}$.

\begin{figure}
    \centering
    \includegraphics[width=0.95\linewidth]{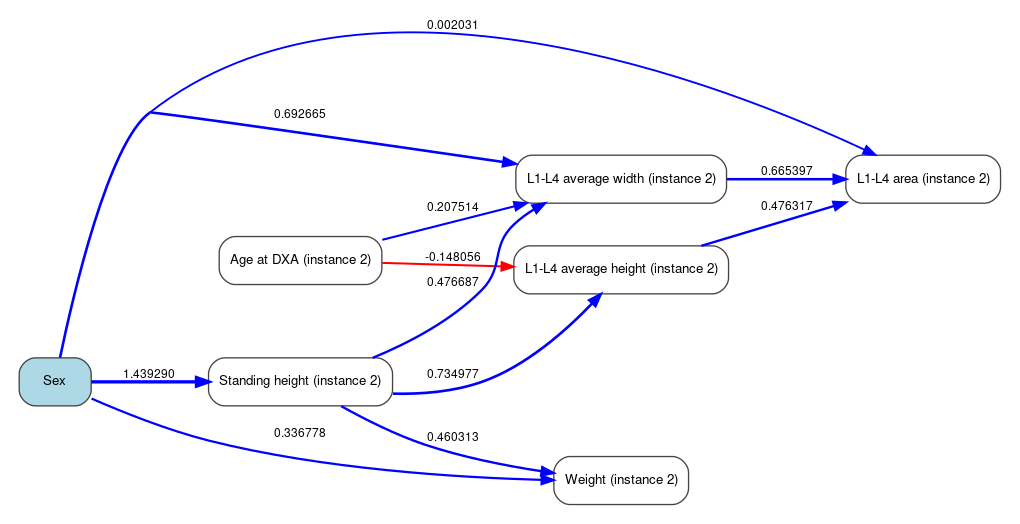}
    \caption{Discovered causal DAG over 3,743 UKB DXA instance 2 (first imaging visit) participants. Edge labels denote standardized local regression coefficients; blue/red indicate positive/negative effects.}
    \label{fig:causal_graph}
\end{figure}

PC-stable \cite{colombo2014order} is applied with mixed-type conditional-independence tests (Fisher’s $Z$ for all-continuous tests and $G^2$ otherwise \cite{tsagris2018constraint}) at $\alpha=0.05$ with Benjamini--Hochberg false discovery rate (FDR) control \cite{benjamini1995controlling}, yielding a completed partially directed acyclic graph (CPDAG) that is oriented via collider detection and Meek’s rules \cite{meek2013causal}. Remaining undirected adjacencies between continuous variables are further oriented using DirectLiNGAM \cite{shimizu2011directlingam} without adding new edges. The resulting DAG (Fig.~\ref{fig:causal_graph}) serves as the explicit SCM for counterfactual generation, edges are optionally labeled with standardized local regression effects for population-level interpretability. This DAG is used as a plausible causal scaffold informed by domain priors and observational data, rather than a definitive causal graph.

\subsubsection{\textbf{Causal Hierarchical Variational Autoencoder (CHVAE)}}
\label{sec:chvae}

A CHVAE jointly models AP spine DXA images and structured covariates under an explicit structural causal model (SCM), following the Deep Structural Causal Model (DSCM) paradigm for tractable counterfactual inference in high-dimensional observations \cite{pawlowski2020deep}. Compared with the standard DSCM using a single-layer latent image model, a hierarchical observation model is introduced, yielding a two-level HVAE ($z_2 \rightarrow z_1 \rightarrow x$) to better capture the multi-scale nature of DXA images. 

In practice, the hierarchy helps decouple global structure from local residual detail, enabling smoother variation under $\mathrm{do}(\mathrm{age})$ sweeps. Similar hierarchical latent designs have been shown to better capture multi-scale structure in images and to improve sample quality and training stability compared with shallow latent formulations \cite{sonderby2016ladder, vahdat2020nvae}.

\textbf{SCM factorization for structured covariates.}
Let $c=(c_1,\dots,c_m)$ denote structured covariates following the discovered directed acyclic graph (DAG), including sex, age, standing\_height, weight, and spine morphometry including l14\_width, l14\_height, l14\_area. Each covariate node $c_j$ is represented as a structural assignment
\begin{equation}
c_j := f_j\big(\mathrm{Pa}(c_j), u_j\big),
\end{equation}
where $\mathrm{Pa}(c_j)$ denotes the parents of node $c_j$ in the DAG and $u_j$ is an exogenous noise variable. 
In our implementation, continuous positive variables are parameterized by sampling base noise from a simple distribution (Normal) in an unconstrained space, then mapping to the observation domain through monotone, (approximately) invertible transforms. Specifically, the root node age is sampled as
\begin{equation}
u_{\text{age}} \sim \mathcal{N}(0,I), 
\qquad 
age = g_{\text{age}}(u_{\text{age}}),
\end{equation}
where $g_{\text{age}}(\cdot)$ is a sequence of invertible transforms (a flow in an unconstrained space followed by a positive-support constraint transform). For non-root nodes, we use conditional affine transforms whose location and scale are predicted from parents (in an unconstrained space), followed by the same positive-support constraint. This yields an explicit SCM in which exogenous variables $\{u_j\}$ can be approximately recovered by inverting the monotone transforms given observed $(\mathrm{Pa}(c_j),c_j)$, supporting counterfactual simulation.

\textbf{Covariate-to-image conditioning.}
The image generator does not take all covariates as direct inputs, instead, it is conditioned on a compact covariate context vector
\begin{equation}
\mathrm{ctx} := \big(l14\_w,\,l14\_h,\,l14\_a\big),
\end{equation}
with,
\begin{equation}
\begin{aligned}
l14\_w &= g^{-1}_{w}(l14\_\mathrm{width}),\\
l14\_h &= g^{-1}_{h}(l14\_\mathrm{height}),\\
l14\_a &= g^{-1}_{a}(l14\_\mathrm{area}).
\end{aligned}
\end{equation}
where  \(g^{-1}\) denotes the inverse of the positive-support constraint transform, l14\_width, l14\_height, and l14\_area denote the L1--L4 morphometry measurements of \emph{width}, \emph{height}, and \emph{area}, respectively. This design injects morphometry-relevant information into the image generator while keeping the full causal dependencies between all covariates encoded in the SCM.

\textbf{Hierarchical latent image mechanism (HVAE).}
Given $\mathrm{ctx}$, the image generator is parameterized by a two-level latent hierarchy $z_2 \rightarrow z_1 \rightarrow x$:
\begin{align}
p(z_2) &= \mathcal{N}(0, I), \\
p_\theta(z_1 \mid z_2) &= \mathcal{N}\!\Big(\mu_{\theta}^{(p)}(z_2), \mathrm{diag}\big(\sigma_{\theta}^{(p)}(z_2)^2\big)\Big), \\
p_\theta(x \mid z_1, \mathrm{ctx}) &= p_\theta\!\Big(x \mid \mathrm{Dec}_\theta([z_1,\mathrm{ctx}])\Big).
\end{align}
In our implementation, $\mu_{\theta}^{(p)}(\cdot)$ and $\sigma_{\theta}^{(p)}(\cdot)$ are produced by a small multilayer perceptron (MLP) head applied to $z_2$, and the decoder $\mathrm{Dec}_\theta(\cdot)$ is a convolutional network with nearest-neighbour upsampling and residual blocks. The likelihood $p_\theta(x \mid z_1,\mathrm{ctx})$ is instantiated as a transformed Laplace distribution to encourage sharper reconstructions, with its scale initialized from a fixed log-scale and capped to avoid degenerate uncertainty and overly-smoothed samples.

\textbf{Hierarchical latents for DXA.}
AP spine DXA images contain variability at different spatial scales: global factors (overall body habitus and coarse spine geometry) affect large regions, whereas local factors (vertebral edges, trabecular texture, and subtle shape differences) are fine-grained. A hierarchical latent structure provides an inductive bias that allocates representational capacity across scales: the top latent $z_2$ captures global, slowly-varying structure, while the lower latent $z_1$ captures finer residual variability conditioned on $z_2$. This improves expressivity compared to a shallow latent model and is beneficial for subject-specific counterfactual generation, where global identity should be preserved while intervention effects propagate through the SCM and the conditional generator.

\textbf{Variational inference.}
Given an observed image $x$ and covariates $c$, an amortized encoder produces a feature representation $h=\mathrm{Enc}(x)$. We concatenate $h$ with the context vector $\mathrm{ctx}$ to form the inference input. The approximate posterior follows the hierarchical factorization:
{\footnotesize
\begin{align}
q_\phi(z_2 \mid x,\mathrm{ctx}) &= \mathcal{N}\!\Big(\mu_{\phi}^{(2)}([h,\mathrm{ctx}]), \mathrm{diag}\big(\sigma_{\phi}^{(2)}([h,\mathrm{ctx}])^2\big)\Big), \\
q_\phi(z_1 \mid x,\mathrm{ctx},z_2) &= \mathcal{N}\!\Big(\mu_{\phi}^{(1)}([h,\mathrm{ctx},z_2]), \mathrm{diag}\big(\sigma_{\phi}^{(1)}([h,\mathrm{ctx},z_2])^2\big)\Big).
\end{align}
}

Training maximizes an evidence lower bound (ELBO) with multiple stochastic variational inference (SVI) particles. We apply KL warm-up by scaling the KL terms with a factor $\beta$ that increases during early epochs:
{\small
\begin{align}
\mathcal{L}
&= \mathbb{E}_{q_\phi}\!\left[\log p_\theta(x\mid z_1,\mathrm{ctx})\right]
- \beta \,\mathrm{KL}\!\left(q_\phi(z_2\mid x,\mathrm{ctx})\,\|\,p(z_2)\right) \nonumber\\
&\quad - \beta \,\mathbb{E}_{q_\phi(z_2\mid x,\mathrm{ctx})}
\left[\mathrm{KL}\!\left(q_\phi(z_1\mid x,\mathrm{ctx},z_2)\,\|\,p_\theta(z_1\mid z_2)\right)\right].
\end{align}
}
In practice, $\beta$ is implemented as a schedule updated per epoch, and the HVAE log-standard deviations are clamped for numerical stability.

\textbf{Counterfactual image synthesis (AAP).}
Counterfactual DXA images under interventions such as $\mathrm{do}(age=\tilde{a})$ are generated via AAP. 

(i) \emph{Abduction}: infer the hierarchical latents by sampling from the variational posterior $q_\phi(z_2,z_1 \mid x,\mathrm{ctx})$ and recover exogenous bases $\{u_j\}$ by inverting the SCM transforms given observed $(\mathrm{Pa}(c_j),c_j)$. 

(ii) \emph{Action}: apply the intervention by replacing the targeted observed node value(s) (e.g. age) while holding the remaining exogenous variables fixed, ensuring the same individual-specific noise realization. 

(iii) \emph{Prediction}: simulate the SCM forward to obtain counterfactual covariates $\tilde{c}$, recompute the corresponding context $\widetilde{\mathrm{ctx}}$, and decode a counterfactual image $\tilde{x}\sim p_\theta(x\mid z_1,\widetilde{\mathrm{ctx}})$ using the same abducted latent/exogenous realization. 

This produces intervention-aligned counterfactual images that preserve subject identity while reflecting the causal effect of the imposed do-operation through the learned SCM and the hierarchical image generator.

\section{Experiments and Results}
\subsection{Implementation details}
\label{sec:impl_details}

We implemented the proposed CHVAE in PyTorch with
Pyro for stochastic variational inference. AP spine DXA images were resized to $192{\times}192$ and normalized to $[0,1]$. Structured covariates (sex, age, standing\_height, weight, l14\_width/height/area) were modeled by an explicit SCM with base Normal noise and monotone constraint transforms (conditional affine transforms for non-root nodes). The two-level HVAE ($z_2{\rightarrow}z_1{\rightarrow}x$) was conditioned on $\mathrm{ctx}=(l14\_w,l14\_h,l14\_a)$ obtained by inverting the constraint transforms. We trained for 1600 epochs by maximizing the ELBO with KL warm-up ($\beta$: $10^{-6}{\rightarrow}1$ over 900 epochs) and clamped log-stds for stability. Counterfactuals were generated via AAP by reusing the abducted latent/exogenous realization under $\mathrm{do}(\cdot)$ interventions.

\subsection{Counterfactual generation}
\label{sec:counterfactual_generation}
\subsubsection{Generation quality}
\label{sec:generation_quality}
Factual reconstruction is first evaluated to assess whether the proposed CHVAE can reproduce observed AP spine DXA images. For each test subject, hierarchical latents are inferred from the image and covariates to obtain a reconstruction $\hat{x}$ by decoding with the inferred codes; when using multiple SVI particles, reconstructions are reported as the particle-wise mean. Reconstruction quality is quantified by comparing $\hat{x}$ with the original image $x$ using MAE, RMSE, PSNR, and SSIM, establishing sufficient observation-model expressivity for subsequent counterfactual experiments. Table~\ref{tab:recon_quality_grouped} reports MAE/RMSE (lower is better) and PSNR/SSIM (higher is better).

\begin{table*}[t]
\centering
\caption{Factual reconstruction quality on the test set.}
\label{tab:recon_quality_grouped}
\setlength{\tabcolsep}{6pt}
\renewcommand{\arraystretch}{1.15}
\begin{tabular}{lcc|cc}
\toprule
\multirow{2}{*}{\textbf{Model}} 
& \multicolumn{2}{c|}{\textbf{Pixel error}$\downarrow$} 
& \multicolumn{2}{c}{\textbf{Perceptual similarity}$\uparrow$} \\
\cmidrule(lr){2-3}\cmidrule(lr){4-5}
& \textbf{MAE} & \textbf{RMSE} & \textbf{PSNR (dB)} & \textbf{SSIM} \\
\midrule
CHVAE (factual reconstruction) 
& $0.0509 \pm 0.0062$ 
& $0.0891 \pm 0.0139$
& $21.10 \pm 1.31$
& $0.574 \pm 0.031$ \\
\bottomrule
\end{tabular}
\end{table*}

Reconstruction results indicate that the proposed HVAE observation model can reproduce the overall intensity distribution and coarse anatomical structure of AP spine DXA images, while leaving room for improvement in fine-grained detail. Specifically, reconstructions achieve $\mathrm{MAE}=0.0509\pm0.0062$ and $\mathrm{PSNR}=21.10\pm1.31$ dB (images normalized to $[0,1]$), suggesting moderate pixel-level fidelity. The structural similarity is $\mathrm{SSIM}=0.574\pm0.031$, which is consistent with partial loss of high-frequency content such as vertebral edge sharpness and subtle trabecular texture.

\subsubsection{Causal consistency under $\mathrm{do}(\mathrm{age})$}
\label{sec:causal_consistency}
We verified counterfactual generation follows the intended causal semantics under interventions on age. For a given subject with observed baseline $(x,c)$, we first abduct subject-specific latent variables and SCM exogenous noises, and then answer a sequence of counterfactual queries $\mathrm{do}(\mathrm{age}=a)$ by reusing the same individual-specific noise realization (and the same inferred latent sample) while replacing only the age node with the intervention value. We report three complementary checks.

\textbf{(i) Intervened-node correctness.}
We confirm that the intervened node matches the target value, i.e.\ $\mathrm{age}_{\mathrm{cf}}\approx a$, ensuring that the do-operator is correctly applied at the observable node.

\textbf{(ii) Causal effect along expected pathways.}
We measure whether downstream morphometry variables respond to $\mathrm{do}(\mathrm{age})$ in a manner consistent with the learned SCM. Concretely, we analyze the dependence between the intervention value $a$ and the generated downstream nodes (e.g. l14\_width, l14\_height, and l14\_area) using summary statistics such as linear slope, Pearson correlation, and $R^2$ over an age sweep.

\begin{figure*}[t]
    \centering
    \includegraphics[width=\textwidth]{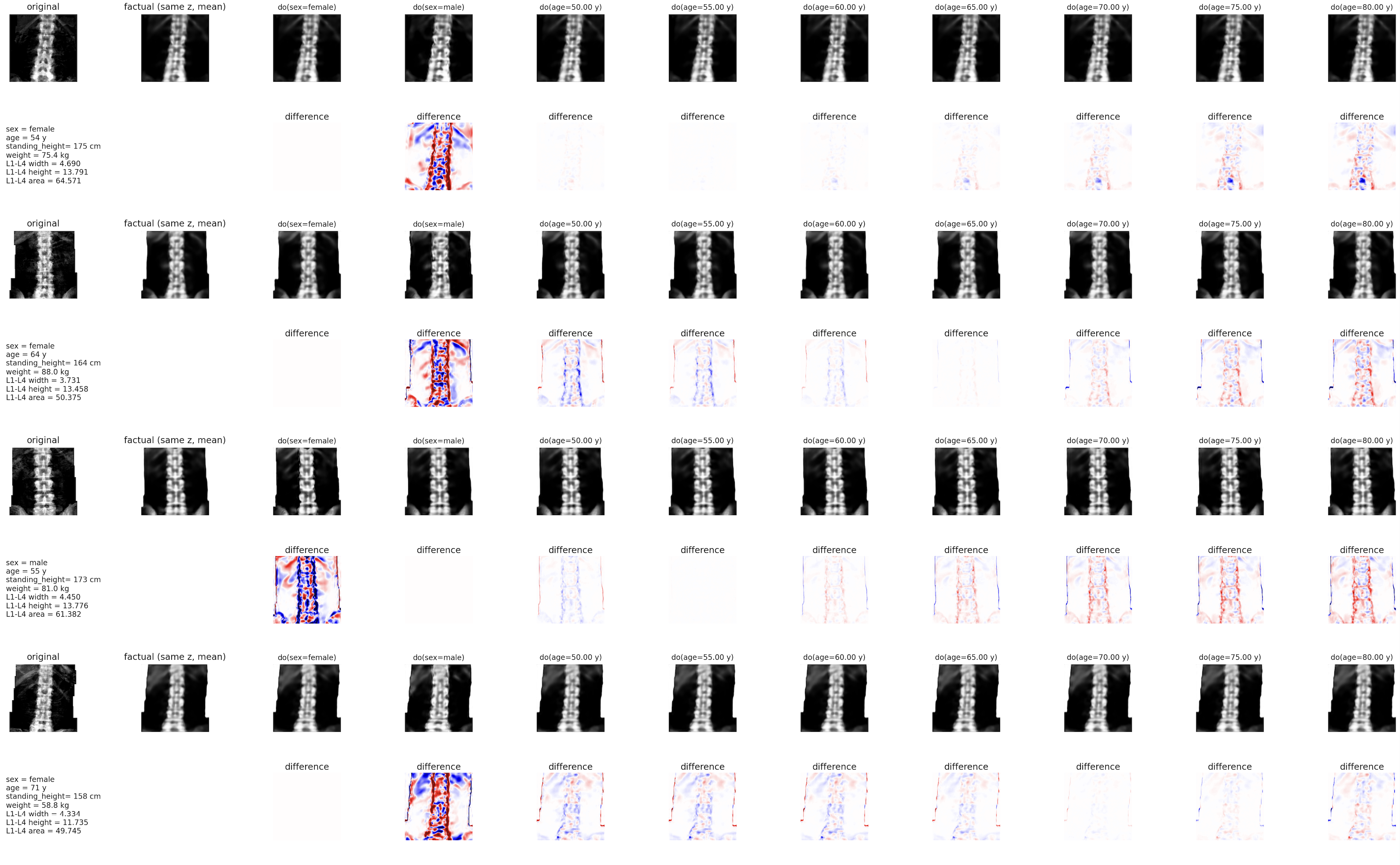}
    \caption{Counterfactual AP spine DXA synthesis under age interventions using the HVAE framework. For each subject, counterfactuals are generated under $\mathrm{do}(\mathrm{age}=a)$ with $a\in\{50,55,\dots,80\}$ by AAP, reusing the same subject-specific exogenous noise. Difference maps are computed w.r.t. factual reconstruction / factual baseline, highlighting localized lumbar (L1--L4) changes.}
    \label{fig:causal_inference}
\end{figure*}

\textbf{(iii) Minimality and invariance.}
Counterfactual images should change only as much as necessary to reflect the intervention while preserving subject identity. Minimality is evaluated by comparing each counterfactual image $x_{\mathrm{cf}}(a)$ with the factual baseline $x_{\mathrm{factual}}$ using MAE/RMSE and SSIM, and invariance is assessed by verifying that non-descendants of age in the SCM remain unchanged under $\mathrm{do}(\mathrm{age})$. The scaling of image-level minimality with intervention magnitude is examined via the relationship between image difference and $|a-a_0|$, where $a_0$ is the factual age. Qualitative examples (factual, counterfactual age sweep, and difference maps) are shown in Fig.~\ref{fig:causal_inference}.

Fig.~\ref{fig:causal_inference} illustrates counterfactual AP spine DXA synthesis under explicit interventions in the learned CHVAE framework. Specifically, we first perform abduction to infer subject-specific exogenous variables and hierarchical latents, then apply a do-intervention on age with $\mathrm{do}(\mathrm{age}=a)$ for $a\in\{50,55,60,65,70,75,80\}$ years (step size 5 years), while keeping non-intervened exogenous noise fixed, and finally predict the downstream covariates and decode the corresponding counterfactual images. The figure shows the original image, its factual reconstruction, and counterfactual samples across the age sweep. Difference maps (counterfactual minus factual) highlight intervention-induced appearance changes, which are primarily localized around the lumbar spine region (L1--L4), consistent with age-driven morphological variation propagating through the SCM to the morphometry-to-image context. The results provide qualitative evidence that the model produces visually plausible and intervention-aligned counterfactual DXA images under age manipulation.

\begin{table*}[!t]
\centering
\caption{Trends of SCM-predicted L1--L4 morphometry under $\mathrm{do}(\mathrm{age})$. For each subject, we report the original (factual) age and L1--L4 morphometry: $W$ denotes the L1--L4 width, $H$ denotes the L1--L4 height, and $A$ denotes the L1--L4 area. Intervention effects are reported as signed differences relative to the original values, $\Delta W(a)=W_{\mathrm{cf}}(a)-W_{\mathrm{orig}}$ (and similarly for $\Delta H(a)$ and $\Delta A(a)$), for $a\in\{50,55,\dots,80\}$.}
\label{tab:do_age_trends_morph_only}
\setlength{\tabcolsep}{2.0pt}
\renewcommand{\arraystretch}{1.15}
\scriptsize
\resizebox{\textwidth}{!}{
\begin{tabular}{c r *{3}{r} *{7}{r} *{7}{r} *{7}{r}}
\toprule
\multicolumn{1}{c}{} &
\multicolumn{4}{c}{Original (factual)} &
\multicolumn{7}{c}{$\Delta W$ at $\mathrm{do}(\mathrm{age}=a)$} &
\multicolumn{7}{c}{$\Delta H$ at $\mathrm{do}(\mathrm{age}=a)$} &
\multicolumn{7}{c}{$\Delta A$ at $\mathrm{do}(\mathrm{age}=a)$} \\
\cmidrule(lr){2-5}\cmidrule(lr){6-12}\cmidrule(lr){13-19}\cmidrule(lr){20-26}
ID & $\mathrm{age}_{\mathrm{orig}}$ & $W_{\mathrm{orig}}$ & $H_{\mathrm{orig}}$ & $A_{\mathrm{orig}}$ &
50 & 55 & 60 & 65 & 70 & 75 & 80 &
50 & 55 & 60 & 65 & 70 & 75 & 80 &
50 & 55 & 60 & 65 & 70 & 75 & 80 \\
\midrule
1 & 54.06 & 4.690 & 13.791 & 64.571 &
$-0.003$ & $+0.001$ & $+0.004$ & $+0.018$ & $+0.032$ & $+0.046$ & $+0.059$ &
$+0.054$ & $-0.011$ & $-0.071$ & $-0.126$ & $-0.215$ & $-0.387$ & $-0.547$ &
$+0.272$ & $-0.058$ & $-0.264$ & $-0.324$ & $-0.544$ & $-1.145$ & $-1.700$ \\
2 & 63.86 & 3.731 & 13.458 & 50.375 &
$-0.102$ & $-0.066$ & $-0.030$ & $+0.008$ & $+0.043$ & $+0.075$ & $+0.106$ &
$+0.182$ & $+0.111$ & $+0.046$ & $-0.014$ & $-0.176$ & $-0.329$ & $-0.470$ &
$-0.731$ & $-0.503$ & $-0.232$ & $+0.066$ & $-0.049$ & $-0.169$ & $-0.286$ \\
3 & 55.15 & 4.450 & 13.776 & 61.382 &
$-0.043$ & $-0.001$ & $+0.037$ & $+0.073$ & $+0.106$ & $+0.137$ & $+0.173$ &
$+0.069$ & $+0.002$ & $-0.059$ & $-0.116$ & $-0.220$ & $-0.388$ & $-0.544$ &
$-0.307$ & $-0.009$ & $+0.262$ & $+0.511$ & $+0.507$ & $+0.180$ & $-0.021$ \\
4 & 71.45 & 4.334 & 11.735 & 49.745 &
$-0.069$ & $-0.061$ & $-0.045$ & $-0.025$ & $-0.005$ & $+0.013$ & $+0.030$ &
$+0.425$ & $+0.375$ & $+0.329$ & $+0.232$ & $+0.050$ & $-0.117$ & $-0.274$ &
$+0.962$ & $+0.844$ & $+0.846$ & $+0.684$ & $+0.148$ & $-0.349$ & $-0.813$ \\

\bottomrule
\end{tabular}
}
\end{table*}

Table~\ref{tab:do_age_trends_morph_only} summarizes subject-wise trends of SCM-predicted L1--L4 morphometry under $\mathrm{do}(\mathrm{age})$ for the first four subjects in Fig.~\ref{fig:causal_inference}, providing a quantitative check of causal consistency. As the intervened age increases from 50 to 80 (5-year steps), the predicted changes in $W$, $H$, and $A$ vary smoothly within each subject, while remaining individual-specific across subjects, consistent with reusing the same abducted exogenous realization and replacing only the age value. The responses also align with the SCM structure: width tends to increase slightly with age whereas height decreases, leading to area changes determined by their balance. Overall, these trends support that the learned SCM implements the intended intervention semantics and propagates effects along the age$\rightarrow$morphometry pathways.

\subsection{Follow-up evaluation}
To assess whether the learned causal model supports longitudinal prediction, a baseline-to-follow-up experiment was conducted using 319 test-set participants with AP spine DXA acquired at imaging baseline (instance~2) and follow-up (repeat scans, recorded as instance~3 in UKB). For each eligible participant, instance~2 images were treated as factual observations, and counterfactual follow-up predictions were generated via abduction--action--prediction (AAP) under $do(\mathrm{age}=\mathrm{age}_{\mathrm{inst3}})$ while reusing abducted identity-related factors.

Counterfactual predictions were compared against the observed instance~3 metadata to evaluate whether the age intervention yields plausible downstream propagation along the structural causal pathways. Because true longitudinal changes in vertebral morphology are expected to be subtle over the follow-up interval, evaluation focused on vertebral shape variables (L1--L4 average width and height, and the derived L1--L4 area). Agreement was assessed at two levels. Absolute-level agreement compares the counterfactual follow-up value $y_{\mathrm{CF}}$ against the observed follow-up value $y_{\mathrm{inst3}}$. Change-level agreement compares the predicted longitudinal change $\Delta_{\mathrm{pred}} = y_{\mathrm{CF}} - y_{\mathrm{inst2}}$ against the observed change $\Delta_{\mathrm{real}} = y_{\mathrm{inst3}} - y_{\mathrm{inst2}}$.

\begin{figure*}[t]
    \centering
    \begin{subfigure}[c]{0.33\linewidth}
        \centering
        \includegraphics[height=4.2cm, width=\linewidth, keepaspectratio]{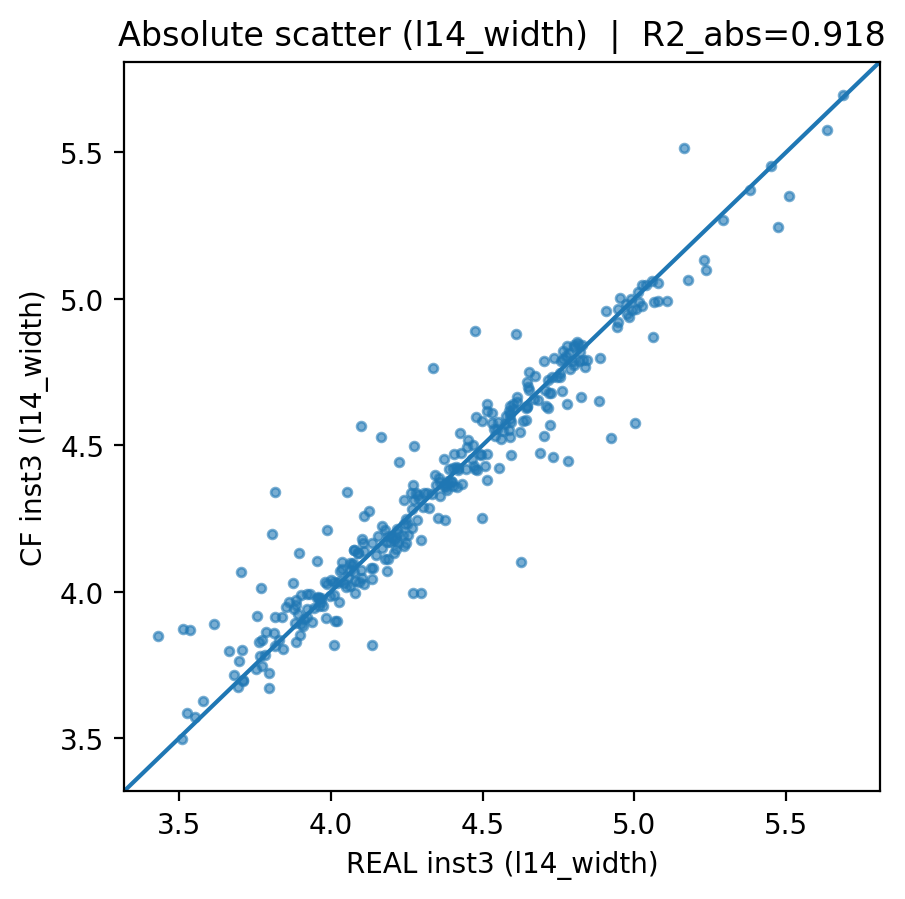}
        \caption{L1--L4 Width: CF vs Real (inst3)}
        \label{fig:followup_abs_width}
    \end{subfigure}\hfill
    \begin{subfigure}[c]{0.33\linewidth}
        \centering
        \includegraphics[height=4.2cm, width=\linewidth, keepaspectratio]{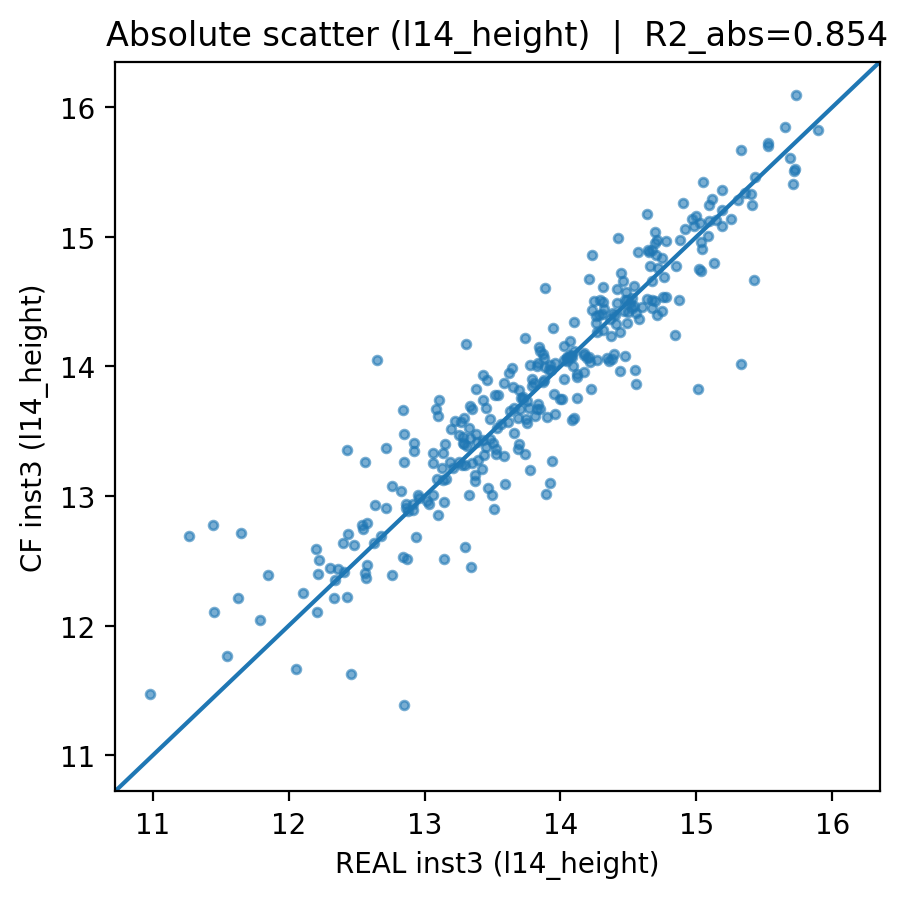}
        \caption{L1--L4 Height: CF vs Real (inst3)}
        \label{fig:followup_abs_height}
    \end{subfigure}\hfill
    \begin{subfigure}[c]{0.33\linewidth}
        \centering
        \includegraphics[height=4.2cm, width=\linewidth, keepaspectratio]{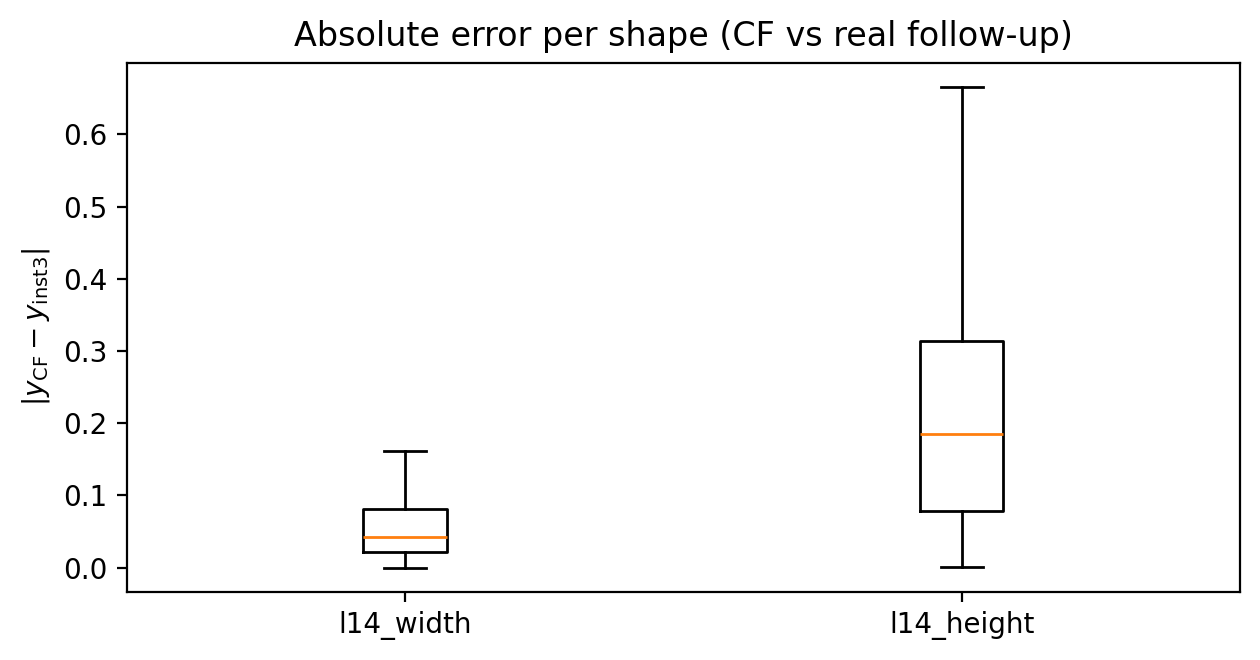}
        \caption{Absolute Error (Width and Height)}
        \label{fig:followup_abs_box_wh}
    \end{subfigure}

    \caption{Follow-up evaluation on absolute L1--L4 shape measurements (counterfactual inst3 vs real inst3).}
    \label{fig:followup_abs_wh_triplet}
\end{figure*}

\paragraph{\textbf{Direct descendants of age: L1--L4 width and height}}
As direct children of age in the discovered causal graph, L1--L4 average width and height exhibit strong absolute-level agreement between counterfactual follow-up predictions and observed instance~3 measurements. In the absolute scatter plots (Fig.~\ref{fig:followup_abs_width}--\ref{fig:followup_abs_height}), predictions closely follow the identity line with $R^2_{\mathrm{abs}}=0.918$ for width and $0.854$ for height, indicating accurate calibration of follow-up levels and preservation of between-subject ordering under the age intervention. The absolute-error distribution (Fig.~\ref{fig:followup_abs_box_wh}) shows low typical errors, with a tighter spread for width than for height, the larger dispersion in height likely reflects additional non-age sources of variability (e.g., posture-related variability, degenerative changes, or residual acquisition noise).

\begin{figure*}[t]
    \centering
    \begin{minipage}[c]{0.25\linewidth}
        \centering
        \includegraphics[width=\linewidth]{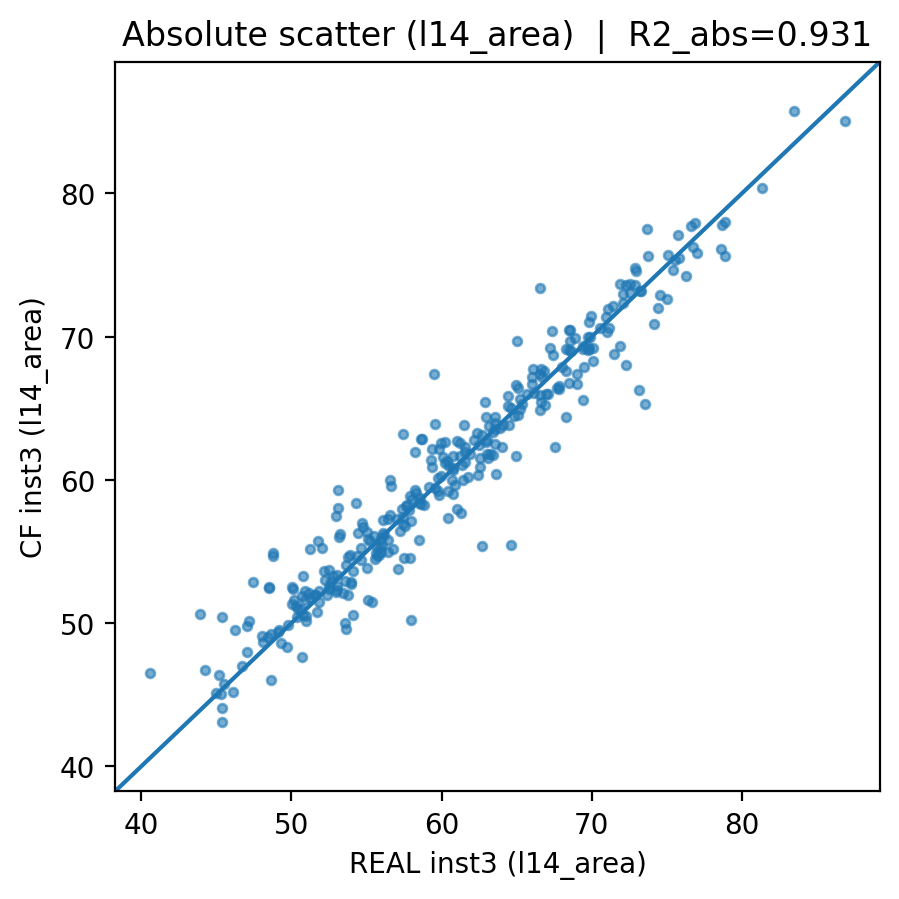}
        \subcaption{L1--L4 Area: CF vs Real (inst3).}
        \label{fig:followup-area-scatter}
    \end{minipage}\hfill
    \begin{minipage}[c]{0.32\linewidth}
        \centering
        \includegraphics[width=\linewidth]{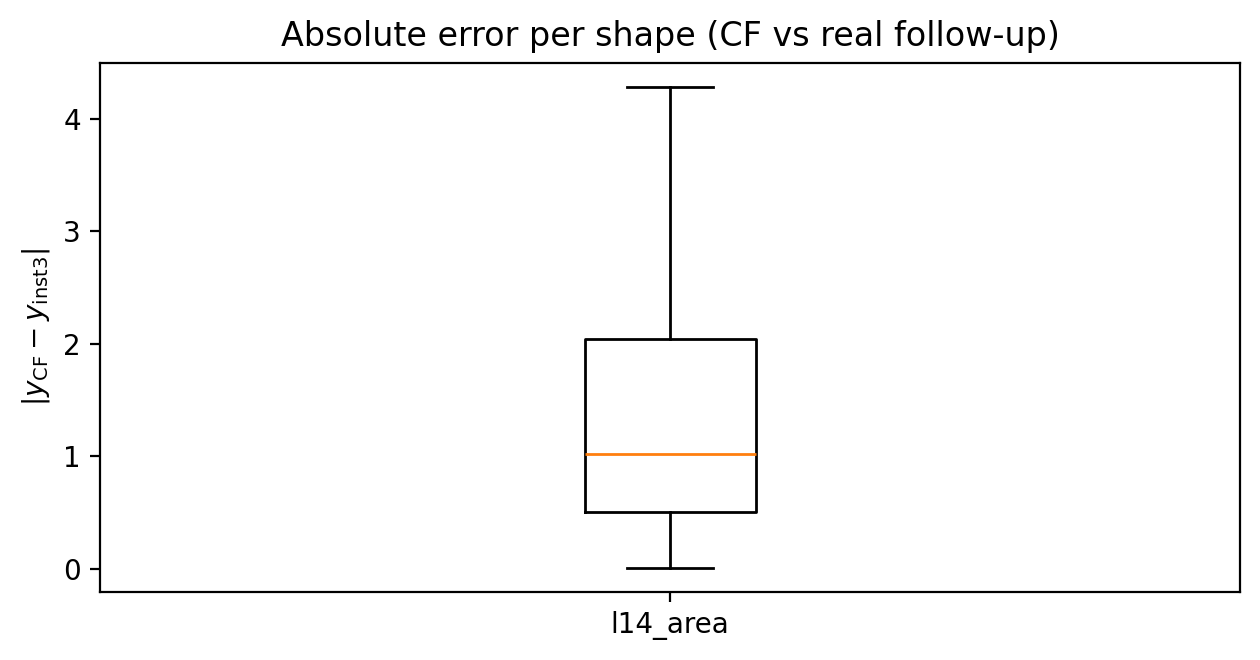}
        \subcaption{Absolute Error Distribution (Area).}
        \label{fig:followup-area-absbox}
    \end{minipage}\hfill
    \begin{minipage}[c]{0.30\linewidth}
        \centering
        \includegraphics[width=\linewidth]{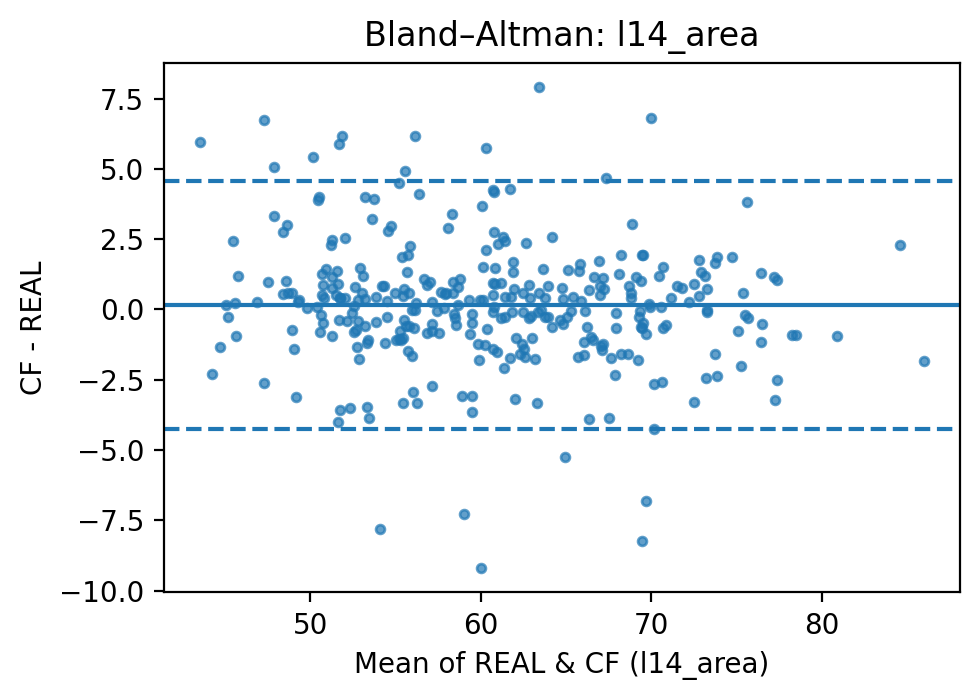}
        \subcaption{Bland--Altman Plot (Area).}
        \label{fig:followup-area-ba}
    \end{minipage}
    \caption{Evaluation of absolute L1--L4 area at follow-up (inst3) using counterfactual predictions generated from baseline (inst2) under $\mathrm{do}(\mathrm{age}=\mathrm{age}_{\mathrm{inst3}})$.}
    \label{fig:followup-area}
\end{figure*}

\paragraph{\textbf{Indirect descendant of age: L1--L4 area}}
L1--L4 area is an indirect descendant of age in the causal graph and therefore constitutes a more challenging target due to the propagation and compounding of upstream uncertainty. Nevertheless, strong absolute-level agreement is observed (Fig.~\ref{fig:followup-area-scatter}) with $R^2_{\mathrm{abs}}=0.931$ and points concentrated around the identity line across the dynamic range. The error distribution exhibits a wider spread and occasional larger deviations (Fig.~\ref{fig:followup-area-absbox}), consistent with area reflecting a composite of multiple shape components. Bland--Altman analysis (Fig.~\ref{fig:followup-area-ba}) indicates minimal systematic bias (mean CF--REAL difference close to zero) but relatively broad limits of agreement, suggesting strong population-level calibration while a subset of individuals shows larger follow-up discrepancies.

\paragraph{\textbf{Directional change agreement}}
Beyond absolute follow-up accuracy, the direction of longitudinal change ($\Delta=\mathrm{inst3}-\mathrm{inst2}$) was assessed using sign agreement. A sign agreement of $0.615$ indicates that 61.5\% of participants have matched change direction between predicted and observed trajectories. This moderate value is expected when true within-subject changes are subtle and $\Delta_{\mathrm{real}}$ is frequently close to zero, such that small prediction errors can flip the sign even if absolute follow-up measurements remain accurate.

Taken together, absolute-level agreement is reported alongside change-level metrics to characterize both calibration and longitudinal trends. Matching $y_{\mathrm{inst3}}$ is a stricter check than matching $\Delta_{\mathrm{real}}$ alone, since $\Delta_{\mathrm{real}}$ can be near zero over short follow-up intervals while the follow-up scale and between-subject ordering must still be preserved. Importantly, $y_{\mathrm{CF}}$ is obtained via counterfactual simulation under $\mathrm{do}(\mathrm{age}=\mathrm{age}_{\mathrm{inst3}})$ using abduction--action--prediction, rather than a direct regression fit to instance~3 targets. The resulting $R^2_{\mathrm{abs}}$ values indicate well-calibrated follow-up morphometry at the cohort level under age-driven counterfactual simulation.

\section{Discussion}
The follow-up evaluation indicates that the proposed causal HVAE produces well-calibrated counterfactual follow-up morphometry under an age intervention. Absolute agreement is strongest for age’s direct descendants (L1--L4 width and height), while the indirect target (L1--L4 area) shows wider dispersion, consistent with uncertainty propagation through derived quantities. Change-direction agreement is more modest for subtle instance~2$\rightarrow$instance~3 changes, motivating joint reporting of absolute- and change-level metrics for longitudinal counterfactual assessment.

This behaviour is supported by two design choices: (i) explicit morphometry-based conditioning links $\mathrm{do}(\mathrm{age})$ to interpretable, region-level variation rather than relying on uncontrolled texture correlations, and (ii) the hierarchical latent space encourages smooth, coherent image variation under small conditioning shifts. Together, these components enable stable counterfactual synthesis that preserves subject-specific anatomy while reflecting intervention-aligned morphological change.

Several limitations remain. Current evaluation focuses on $\mathrm{do}(\mathrm{age})$ and morphometry endpoints, leaving causal consistency under other covariates and clinically grounded outcomes (e.g., BMD/T-score) unexplored. In addition, training on heterogeneous raw DXA may yield over-smoothed outputs, and the present follow-up setup uses single-variable interventions without explicitly modelling multi-factor longitudinal processes or potential unmeasured confounding.

Future work will extend the intervention space to additional covariates and multi-variable scenarios, link counterfactual images to clinical endpoints (BMD/T-score and osteoporosis-related analyzes), incorporate genetic features to capture heterogeneous trajectories, and explore stronger conditional generators with improved uncertainty calibration to better represent subtle, localized changes.

\section{Conclusion}
This work explored causal-consistent counterfactual generation of AP spine DXA images using a metadata-conditioned hierarchical variational autoencoder trained on 3,743 UKB scans. By embedding structured covariates within an explicit intervention framework, the model supports abduction--action--prediction queries and produces anatomically plausible samples under controlled changes in age. Across the baseline-to-follow-up evaluation, absolute agreement on key vertebral morphometry measurements suggests that the learned conditional generative mapping preserves between-subject ordering while allowing intervention-aligned variation. These results provide evidence that causal conditioning can serve as a practical foundation for longitudinal, scenario-based DXA synthesis at the cohort scale, bridging high-dimensional image generation with interpretable intervention semantics. Future work will connect counterfactual images to clinically grounded endpoints and extend the intervention space to richer covariate sets and more realistic multi-variable scenarios.

\section*{Ethics Statement}
The experimental procedures involving human participants were approved as part of the UK Biobank study by the North West Multi-centre Research Ethics Committee. All participants provided informed consent, and this study used de-identified UK Biobank data under Application Number 700191.

% \section*{References}
\bibliographystyle{IEEEtran}
\bibliography{references}
\end{document}